%% file: templateArxiv.tex
\documentclass{article}

\usepackage{placeins}

\usepackage{PRIMEarxiv}
\usepackage{caption} 
\usepackage{graphicx} 
\usepackage{float}  
\usepackage{placeins} 
\usepackage{multirow}
\usepackage[utf8]{inputenc} 
\usepackage[T1]{fontenc}    
\usepackage{hyperref}       
\usepackage{url}            
\usepackage{booktabs}       
\usepackage{amsmath}       
\usepackage{amsfonts}       
\usepackage{nicefrac}       
\usepackage{enumitem}
\usepackage{lipsum}
\usepackage{fancyhdr}       
\usepackage{graphicx}       
\graphicspath{{media/}}     

\title{Beyond Edge Maps: Wavelet-Domain Conditioning for Multi-Adapter Map-to-Satellite Diffusion}

\author{
  Arisha Prasain\\
  Department of Electronics and Computer Engineering \\
  Pulchowk Engineering Campus, Tribhuvan University \\
  Lalitpur, Nepal\\
  \texttt{arishaprasain0@gmail.com} \\
}
\usepackage{pgffor} 

\begin{document}
\maketitle

\begin{abstract}
\input{abstract}
\end{abstract}

\section{Introduction}
\input{introduction}

\section{Related Works}
\input{relatedWork}

\section{Methodology}
\input{methodology}

\section{Dataset}
\input{dataset}

\section{Experiments}
\input{experiments}

\FloatBarrier
\section{Conclusion}
\input{conclusion}

\bibliographystyle{unsrt}  
\bibliography{references}

\end{document}

%% file: abstract.tex
Commercial mapping partnerships are often unavailable in low-resource
regions, leaving satellite basemaps stale and motivating synthesis of
satellite imagery from independently maintained cartographic data.
Existing ControlNet-based diffusion methods typically condition on
structural signals like edges or segmentation extracted from the
target image itself, assuming the imagery already exists and limiting
their use exactly where synthesis matters most. Map-conditioned
alternatives add cues like edge detection but omit frequency-domain
structure. We propose a ControlNet-based diffusion framework
conditioned only on cartographic sources obtainable independently of
the target imagery: OpenStreetMap (OSM) raster maps and their
stationary wavelet transform (SWT) subbands, a conditioning signal
previously unexplored for map-to-satellite diffusion. Two ControlNet
adapters, trained separately on the map and wavelet representations
atop a frozen Stable Diffusion backbone, are fused via MultiControlNet,
jointly drawing on spatial structure and frequency detail without
retraining a multi-input model. We evaluate on a new paired
map-satellite dataset curated for Nepal, a data-scarce, topographically
diverse region, alongside the Pix2Pix maps-satellite benchmark.
Combined conditioning wins six of eight metric-dataset comparisons --
SSIM and PSNR on both datasets, plus LPIPS (both Alex and VGG backbones) on ours 
and ties map-only on both Pix2Pix LPIPS backbones while still edging
past wavelet-only there. Wavelet-only takes the lowest FID on both
datasets, matching the tradeoff between per-image fidelity and
distributional realism. We treat this gap cautiously given our modest
test-set sizes and FID's known small-sample bias.

\keywords{Satellite-to-Map Translation \and Diffusion Models \and ControlNet \and Remote Sensing \and Nepal \and Multi-Conditional Generation \and Image-to-Image Translation}

%% file: introduction.tex
Satellite imagery supports a wide range of applications, including urban planning, disaster response, environmental
monitoring, agricultural assessment, and infrastructure development. These applications depend on the availability
of recent, high-resolution imagery for the region of interest. However, high-resolution satellite imagery is costly
to collect and infrequently updated, so imagery in developing regions is often not refreshed in step with changes in
the built environment \cite{liu2023building}. Nepal illustrates this disparity: its topographically complex and economically constrained
landscape has historically attracted limited investment in high-frequency commercial satellite revisits. Map data for
such regions, by contrast, can be maintained through channels independent of commercial satellite providers, including
community contributions, GPS-based traces, and government survey integration. This gap between map data currency and
satellite imagery availability raises a central question: can satellite imagery be synthesized directly from cartographic
data, offering a scalable complement to direct acquisition in regions where it is most needed?

Prior work on map-to-satellite synthesis moved from GAN-based translation \cite{isola2017image}
toward diffusion models steered by ControlNet-style conditioning on OpenStreetMap data
\cite{espinosa2023generate,sastry2024geosynth}, with each step bringing gains in
realism and control over the generated layout. Two limitations persist across this line of work,
though. Methods that condition on signals extracted from the target satellite image itself reach
strong pixel-level agreement \cite{tang2024crsdiff}, but that signal doesn't exist in the deployment settings this work
targets, where no recent imagery is available to extract it from in the first place. Methods that
rely only on independently obtainable inputs, such as OSM layouts, sidestep that problem but stay
confined to the spatial domain, leaving the frequency-domain structure already present in the map
itself unused as a conditioning signal. We review this progression and its gaps in detail in
Section 2.

To address this gap, we propose a ControlNet-based diffusion framework that conditions satellite image synthesis exclusively on cartographic sources obtainable independently of target imagery: OpenStreetMap-derived raster maps and their stationary wavelet transform (SWT) subbands. We use SWT rather than the standard discrete wavelet transform because of its shift-invariance, which avoids spatial misalignment artifacts that downsampling-based transforms can introduce when used as a per-pixel conditioning signal \cite{nason1995swt}. The decomposition's directional subbands (LH, HL, HH) capture horizontal, vertical, and diagonal edge and texture information across multiple scales, giving a frequency-domain complement to the purely spatial structure already captured by the OSM raster. Rather than retraining a single model to jointly ingest both modalities, we train two ControlNet adapters independently atop a frozen Stable Diffusion backbone — one on map rasters, one on wavelet subbands and combine them through intermediate fusion. This keeps each adapter specialized to its respective signal while staying modular: either can be used alone, or fused, without altering the backbone or requiring joint retraining.

%% file: relatedWork.tex
\subsection{Map-to-Satellite and Remote Sensing Image Synthesis}
Early approaches to cartographic-satellite image translation relied primarily on generative adversarial networks. Pix2Pix \cite{isola2017image} established the paired conditional image-to-image translation paradigm underlying much of this line of work, framing map-to-aerial translation as a supervised task built on paired data. CycleGAN \cite{zhu2017unpaired} relaxed this pairing requirement, learning cross-domain translation through cycle-consistency alone given how scarce aligned map-satellite pairs remain in many regions. WGAN \cite{arjovsky2017wasserstein} addressed a more fundamental limitation in the adversarial objective itself, replacing the standard GAN loss with a Wasserstein-distance formulation to improve training stability and reduce mode collapse. Even with these advances, GAN-based methods have generally struggled to preserve fine-grained texture and structural detail, particularly across topographically varied terrain.
Dhariwal and Nichol \cite{dhariwal2021diffusion} showed that diffusion models can surpass GANs in sample fidelity, establishing diffusion as the stronger foundation for high-quality conditional image synthesis. Espinosa and Crowley \cite{espinosa2023generate} were among the first to show that satellite imagery could be generated directly from OpenStreetMap data using a diffusion-based ControlNet, establishing that map layouts alone provide enough structural conditioning for plausible satellite synthesis. At the foundation-model scale, DiffusionSat \cite{khanna2024diffusionsat} shows that a single diffusion backbone can be adapted across multiple satellite image generation and editing tasks, positioning diffusion models as a viable general-purpose foundation for remote sensing imagery. CRS-Diff \cite{tang2024crsdiff} pushes further toward controllability, fusing several conditioning modalities like text, metadata, sketches, segmentation masks, and road maps through a multi-control mechanism to produce more precise, realistic remote sensing imagery.
Taken together, this body of work traces a clear progression from GAN-based to diffusion-based synthesis, with steadily growing emphasis on controllability. Existing methods, however, largely stay within the spatial domain and don't explicitly draw on the multi-scale frequency information already present in map inputs — a gap we return to in Section 2.3.

\subsection{ControlNet-Based Conditioning Strategies for Satellite Synthesis}
GeoSynth \cite{sastry2024geosynth} extends the OSM-conditioned ControlNet approach introduced by
Espinosa and Crowley \cite{espinosa2023generate}, adding textual style prompts and geographic location
embeddings via SatCLIP, and evaluating two further control signals: Canny edge maps and Segment Anything
(SAM) segmentation masks, both extracted directly from the target satellite image. Of these,
Canny-conditioned generation reportedly achieves the highest SSIM, which on its face suggests strong
pixel-level agreement with ground truth. That result is hard to read as evidence of practical utility,
though: Canny edges pulled from the target image already encode substantial structural information
about the very output the model is meant to synthesize, so a high SSIM here reflects near-oracle
structural cues rather than the model's ability to generate accurate imagery from independently
obtainable inputs. Recent satellite imagery simply isn't available in the deployment scenarios this
work targets, meaning Canny and SAM-based conditioning can't be constructed there at all. CRS-Diff
\cite{tang2024crsdiff} takes a broader multimodal approach, fusing text, metadata, sketches,
segmentation masks, and road maps through a unified control mechanism, but its use of segmentation
masks carries the same deployment-validity limitation.

\subsection{Wavelet and Frequency-Domain Representations in Generative Models}

Wavelet representations are a common tool for capturing multi-scale and high-frequency structure in
generative and restoration architectures. Diffusion models in particular have rapidly become a default
choice across remote sensing image generation, enhancement, and interpretation \cite{liu2024diffusionrs},
and wavelets are one of several routes researchers have taken to sharpen the frequency content these
models recover. Wavelet Diffusion Models \cite{phung2023wavediff} embed a wavelet subband decomposition
directly inside the diffusion architecture; this cuts the spatial resolution handled at each denoising
step and speeds up both training and sampling with comparable or improved FID relative to pixel-space baselines on most of their benchmarks. WaveDM
\cite{huang2024wavedm} takes a similar wavelet-domain idea into image restoration, where low-frequency
content is split from high-frequency detail so each can be reconstructed on its own terms. For
super-resolution, quaternion wavelet-conditioned diffusion \cite{sigillo2025quaternion} pairs quaternion
representations with wavelet subbands to steer denoising toward finer texture recovery. A diffusion
transformer conditioned on CLIP image embeddings \cite{zhu2025imagetoimage} takes a different route:
global semantic features, rather than frequency-domain signals, can also drive high-fidelity paired
image translation. Wavelets appear outside diffusion too. WDIG \cite{zhu2023wdig} builds its loss
functions in the wavelet domain for GAN-based generation, with low- and high-frequency components
separated to constrain style transfer, image translation, and GAN inversion. Within remote sensing
specifically, CFRWD-GAN \cite{wei2023cfrwdgan} adds a discrete wavelet decomposition branch to a GAN for
SAR-to-optical translation, which suppresses speckle noise and recovers high-frequency detail the
translation would otherwise lose. C-DiffSET \cite{do2024cdiffset} tackles the closely related SAR-to-EO
problem from the opposite direction: rather than decomposing the signal into wavelet subbands at all, it
fine-tunes a pretrained latent diffusion model.

Diffusion-based map generation from remote sensing imagery has progressed along a separate track that
doesn't touch wavelets at all. MapGen-Diff \cite{tian2024mapgendiff} reframes remote-sensing-to-map
translation as a denoising diffusion bridge with fixed source and target endpoints, while SCGM
\cite{sun2025bridgingscalesmapgeneration} adds a cascading, scale-aware generation scheme so tiles produced at coarser zoom
levels condition the synthesis of finer ones. A related line of work, exemplified by the RSFSG framework
\cite{yuan2023efficient}, prioritizes training and sampling efficiency for diffusion-based remote sensing
sample generation through frequency-aware knowledge distillation, again without treating frequency
decomposition as a conditioning signal for the generative model itself. Across this body of work,
wavelet representations prove effective across diffusion, restoration, and GAN-based generation,
including within remote sensing image translation itself. In every case, though, wavelets are built into
the generative architecture, loss function, or optimization process rather than supplied as an external
structural conditioning signal to a ControlNet, and their use as a ControlNet conditioning input for
map-to-satellite diffusion synthesis remains unexplored.

%% file: methodology.tex
\subsection{ControlNet Adapter Mechanism}
\label{sec:controlnet-mechanism}
\begin{figure}[!ht]
    \centering
    \includegraphics[width=0.8\linewidth]{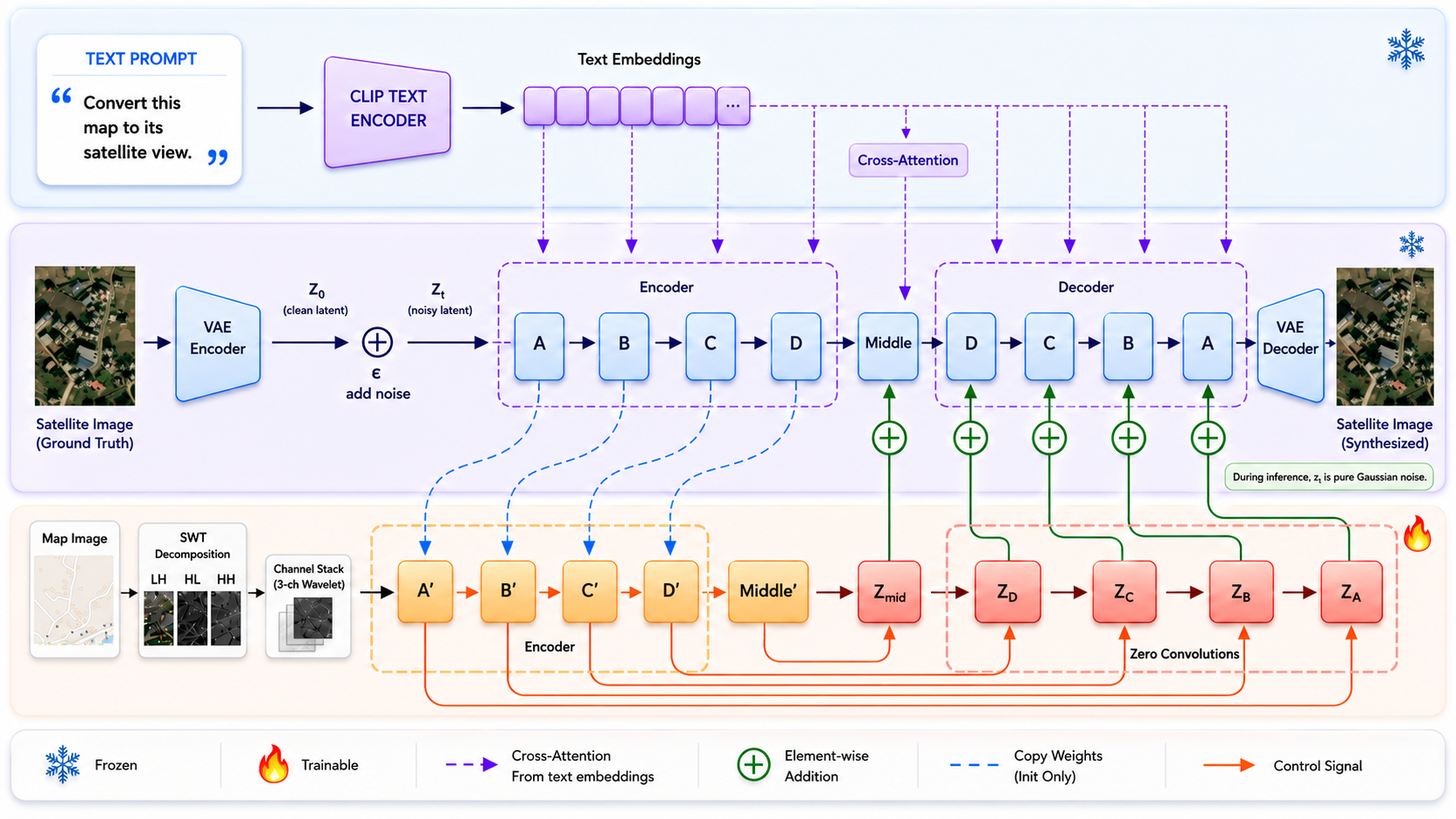}
    \caption{Detailed view of the ControlNet adapter mechanism for a single conditioning branch.}
    \label{fig:controlnet-mechanism}
\end{figure}
\FloatBarrier

Both conditioning signals are injected into the frozen backbone through the
ControlNet paradigm~\cite{zhang2023controlnet}. The two branches described in
Sec.~\ref{sec:branches} share an identical design, so we walk through the
adapter once here, for a single conditioning branch, before turning to the
specific modalities used in this work.

Given a conditioning input, ControlNet builds a trainable copy of the frozen
U-Net's encoding blocks. The conditioning signal first passes through a
lightweight convolutional stem and a zero-initialized convolution, and the
result is merged into the same noisy latent that the frozen backbone
receives, before that combined signal enters the trainable copy. A second
zero-initialized convolution then projects the trainable copy's output before
it is summed back into corresponding skip connection feeding the frozen decoder.
Both connecting layers start at zero, so the adapter contributes nothing to
the pretrained backbone at initialization; this keeps early training stable
and leaves the frozen model's generative prior untouched at the
outset~\cite{zhang2023controlnet}. Fig.~\ref{fig:controlnet-mechanism} shows
this mechanism for a single branch. The OSM branch and the wavelet branch
introduced next each instantiate it independently, differing only in which
conditioning signal feeds the trainable copy.
\subsection{Conditioning Branches}
\label{sec:branches}

Following the mechanism above, two adapters are instantiated, one per
conditioning modality.

\noindent\textbf{OSM Conditioning Branch.}
The first adapter conditions generation on OpenStreetMap raster tiles, giving
the model direct access to cartographic structure. The tile passes through a
lightweight convolutional stem before entering the trainable encoder copy, so
the model can draw on the road networks, building footprints, and land-use
boundaries the map already encodes.

\noindent\textbf{Wavelet Conditioning Branch.}
A second adapter conditions on the Stationary Wavelet Transform (SWT)
of the same OSM tile used by the OSM branch, giving the model a
frequency-domain view of the cartographic content alongside its raw raster
form. Where the standard discrete wavelet transform downsamples the signal at
each level, the SWT instead upsamples the filter coefficients, so it stays
translation-invariant and every subband keeps the tile's full spatial
resolution ~\cite{nason1995swt}. That property matters here: the standard DWT would produce lower-resolution subbands requiring upsampling to match the tile's resolution, and, being shift-variant, would give inconsistent coefficients under the small pixel-level misalignments common between rendered tiles, either of which would break the pixel-wise correspondence the two branches need once summed at every U-Net stage. The tiles are decomposed using a level-2 SWT with a db2
basis~\cite{daubechies1988orthonormal}, which yields an
approximation (LL) and three directional detail subbands (LH, HL,
HH) at each of the two decomposition levels. The LL subbands are
discarded, as they are redundant with the map already supplied to
the OSM branch; the two levels of LH, HL, and HH are then fused by
averaging same-direction subbands across levels, yielding a single
horizontal, vertical, and diagonal-detail map that jointly
capture fine and coarse-scale structure. These three fused
subbands are concatenated into a three-channel input and fed into
this adapter.

\label{sec:fusion}
\begin{figure}[!ht]
    \centering
    \includegraphics[width=0.8\linewidth]{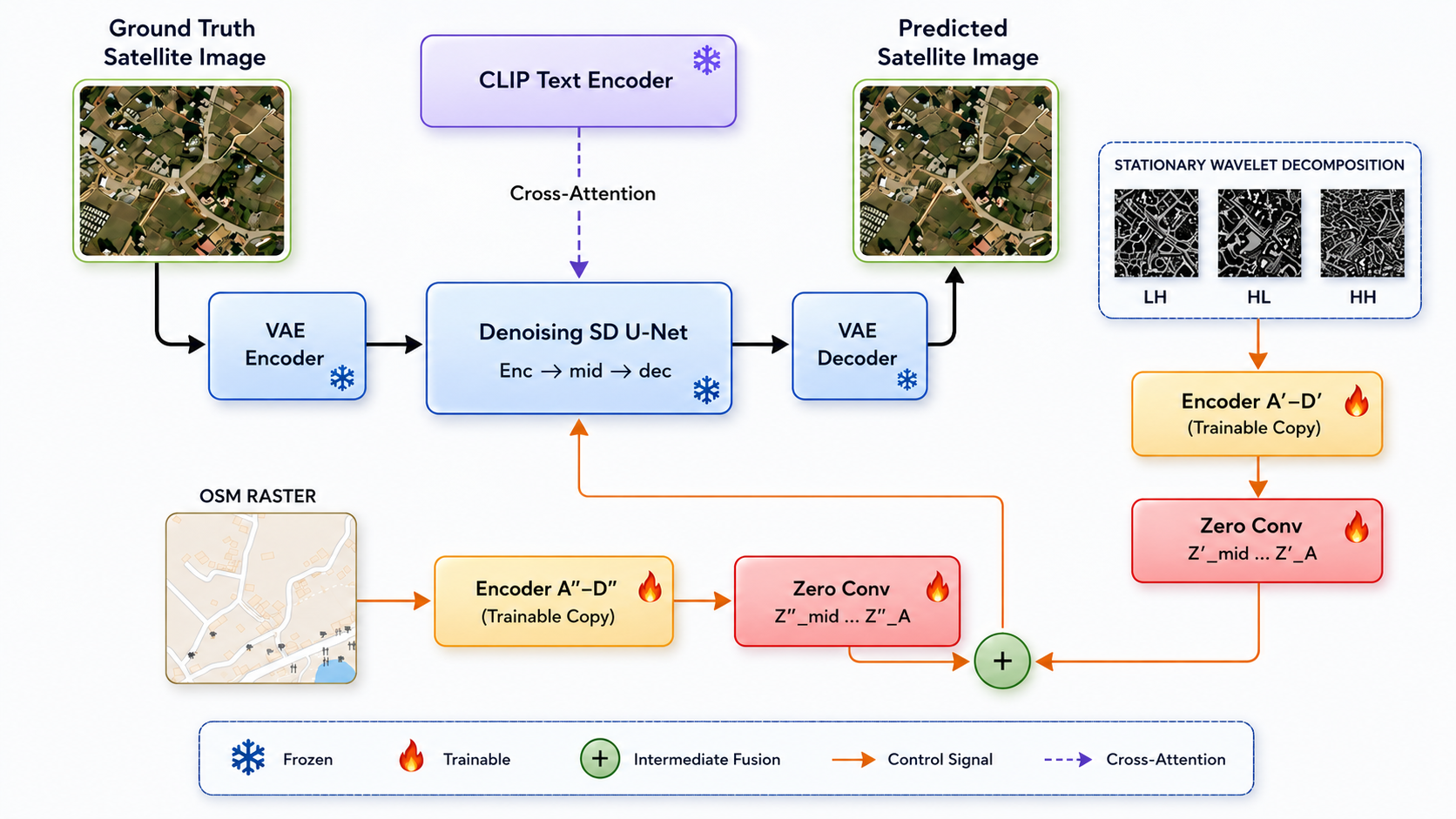}
    \caption{Full dual-adapter architecture. The OSM branch and the
    wavelet branch each instantiate the ControlNet mechanism of
    Fig.~\ref{fig:controlnet-mechanism} independently; their outputs are
    combined at every U-Net encoder stage before being added to the frozen
    decoder.}
    \label{fig:dual-architecture}
\end{figure}

\subsection{Combining the Two Branches}
The OSM and wavelet adapters are combined via intermediate fusion,
implemented as a weighted MultiControlNet composition: each branch's
output passes through its own zero convolution and is scaled by a
per-branch conditioning weight and the two are summed and added to the
corresponding skip connection at each of the U-Net's resolution stages.
Zhang et al.\ introduced composing multiple ControlNets as a direct,
unweighted sum~\cite{zhang2023controlnet}; here, we make use of a per-branch conditioning weight for each branch, which is what lets the ablation in Table ~\ref{tab:ablation} sweep the OSM and wavelet conditioning strengths against each other to study how relative conditioning strength trades cartographic fidelity against frequency-domain detail.

At inference time, a
latent is sampled from a standard normal distribution and denoised over $T$
timesteps, guided jointly by the text embedding, the OSM branch, and the
wavelet branch. Fig.~\ref{fig:dual-architecture} shows the resulting
dual-adapter architecture in compact form; each adapter block follows the
mechanism detailed in Fig.~\ref{fig:controlnet-mechanism}.
\FloatBarrier



%% file: dataset.tex
We introduce a paired map–satellite dataset covering 43 districts across Nepal's hilly and terai regions. Sampling locations came from a shapefile-based approach: for each target district, we built a region boundary either by intersecting a bounding box with Nepal's national boundary shapefile, or, for districts sampled around major urban centers, by extracting the drivable road network from OpenStreetMap and buffering it by 200 meters to define a dense urban sampling zone. Within each boundary we drew candidate points at random, then pooled them across all 43 districts to form the final coordinate set. For every sampled location we rendered a map tile through the Thunderforest vector API and retrieved a co-registered satellite tile from ArcGIS \cite{esri_world_imagery} at zoom level 17.

The dataset comprises 4,885 paired images at 256$\times$256 resolution,
split 80/10/10 into training, validation, and test sets (3,903 / 491 /
491 pairs). Its mix of hilly and terai terrain, and of dense urban zones
alongside broader district-level regions, brings substantial variation
in topography and land cover. Alongside our own data, we also evaluate
on the Pix2Pix dataset~\cite{isola2017image}: 2,194 paired images,
split the same 80/10/10 way into 1,754 training, 220 validation, and
220 test pairs.

%% file: experiments.tex
\subsection{Details}
Our pretrained LDM backbone is Stable Diffusion v2.1~\cite{rombach2022ldm},
built from a Variational Autoencoder (VAE) encoder--decoder pair, a U-Net
denoiser conditioned via cross-attention, and a frozen OpenCLIP text
encoder~\cite{cherti2023reproducible} that supplies the text embedding. An
input image is mapped by the encoder into a compact latent, and the decoder
reconstructs the image from that latent. Every component of the SD backbone
-- the VAE encoder and decoder, the OpenCLIP text encoder, and the U-Net
denoiser stay frozen throughout training; only the two ControlNet
adapters described above are updated. 

All experiments were conducted in a Kaggle environment equipped with dual NVIDIA T4 GPUs (2 × 15 GB VRAM). We trained the ControlNet adapters atop the frozen Stable Diffusion 2.1 for 30,000 training steps on our dataset and 20,000 training steps on the Pix2Pix dataset using the AdamW optimizer with a learning rate of $1 \times 10^{-5}$, a batch size of 4, and the fixed text prompt: "Convert this map to its satellite view." During inference, we used the DDIM sampler~\cite{song2021ddim} with 50 sampling steps and a classifier-free guidance (CFG) scale of 7.5. We evaluate our method on both our proposed dataset and the Pix2Pix dataset~\cite{isola2017image}.

\subsection{Evaluation Metrics}
We evaluate all methods using standard image quality metrics:
\begin{itemize}
    \item \textbf{SSIM} (Structural Similarity Index)~\cite{wang2004ssim} captures how well the generated satellite images preserve the structural patterns of the ground truth, with higher scores indicating better preservation.
    \item \textbf{PSNR} (Peak Signal-to-Noise Ratio), reported in decibels (dB), quantifies pixel-level reconstruction quality; a higher value means the output is closer to the ground truth.
    \item \textbf{FID} (Fr\'echet Inception Distance)~\cite{heusel2017fid} compares the distribution of generated satelite images against real ones, with lower scores corresponding to more realistic generations.
    \item \textbf{LPIPS} (Learned Perceptual Image Patch Similarity, using AlexNet and VGG backbones)~\cite{zhang2018lpips} compares generated and ground truth images in a learned feature space, where a lower distance means the two are perceptually closer.
\end{itemize}

\subsection{Quantitative Results}
Table \ref{tab:results_our} and \ref{tab:results_pix2pix_split} presents the quantitative metrics tested on our dataset and the Pix2Pix dataset respectively.

\begin{table}[h]
\centering
\caption{Quantitative results on our test set of 491 pairs of images.}
\label{tab:results_our}
\begin{tabular}{lccccc}
\hline
\textbf{Variant} &
\textbf{SSIM} $\uparrow$ &
\textbf{PSNR} $\uparrow$ &
\textbf{FID} $\downarrow$ &
\textbf{LPIPS (Alex)} $\downarrow$ &
\textbf{LPIPS (VGG)} $\downarrow$ \\
\hline
OSM Only        & 0.18 & 14.44 & 82.37 &  0.58 & 0.65 \\
Wavelets Only     & 0.20 & 15.42 & 79.23  & 0.57 & 0.64 \\
OSM + Wavelets   & 0.22 & 15.84 & 94.21   & 0.55 & 0.62 \\
\hline
\end{tabular}
\end{table}

\begin{table}[h]
\centering
\caption{Quantitative results on Pix2Pix test set of 220 pairs of images.}
\label{tab:results_pix2pix_split}
\begin{tabular}{lccccc}
\hline
\textbf{Variant} &
\textbf{SSIM} $\uparrow$ &
\textbf{PSNR} $\uparrow$ &
\textbf{FID} $\downarrow$ &
\textbf{LPIPS (Alex)} $\downarrow$ &
\textbf{LPIPS (VGG)} $\downarrow$ \\
\hline
OSM Only        & 0.12 & 12.72 & 124.26 &  0.43 & 0.54 \\
Wavelets Only     & 0.14 & 12.90 & 115.82  & 0.45 & 0.56 \\
OSM + Wavelets   & 0.16 & 14.11 & 120.74   & 0.43 & 0.54 \\
\hline
\end{tabular}
\end{table}

\subsection{Ablation Study}
To evaluate how each condition contributes to generation quality, we
conduct an ablation study by varying the conditioning strength
assigned to the map and wavelet signals with different combinations in the ControlNet input. Results are reported in Table \ref{tab:ablation}.

\begin{table}[h]
\centering
\caption{Ablation study on conditioning scale weights.}
\label{tab:ablation}
\begin{tabular}{lccccccc}
\hline
\textbf{Dataset} &
\textbf{OSM scale} &
\textbf{Wavelets scale} &
\textbf{SSIM} $\uparrow$ &
\textbf{PSNR} $\uparrow$ &
\textbf{FID} $\downarrow$ &

\textbf{LPIPS (Alex)} $\downarrow$ &
\textbf{LPIPS (VGG)} $\downarrow$ \\
\hline
\multirow{3}{*}{Ours}    & 0.5 & 1.0 & 0.22 & 15.84 & 94.21  & 0.55 & 0.62 \\
                         & 1.0 & 0.5 & 0.21 & 15.68 & 96.94   & 0.56 & 0.63 \\
                         & 1.0 & 1.0 & 0.23 & 16.10 & 105.68  & 0.56 & 0.61 \\
\hline
\multirow{3}{*}{Pix2Pix} & 0.5 & 1.0 & 0.160   & 14.11    & 120.74      & 0.43   & 0.54   \\
                         & 1.0 & 0.5 & 0.162   & 14.01    & 123.44      & 0.42   & 0.53   \\
                         & 1.0 & 1.0 & 0.167   & 14.67    & 152.18      & 0.43   & 0.55   \\
\hline
\end{tabular}
\end{table}

\subsection{Discussion}
Wavelet-only conditioning outperforms map-only conditioning on all metrics across our dataset and 3 out of 5 on Pix2Pix. The one exception is Pix2Pix, where map-only outperforms wavelet-only on both LPIPS variants. That points to something specific: the SWT subbands hand the frozen backbone more direct structural cues, things like road edges and boundaries, than the raw OSM raster manages on its own. 

Combining the two signals (with OSM = 0.5, wavelets = 1.0, the best model) yields further gains on the metrics we
weight most heavily. On our own dataset, the combined model achieves
the best SSIM, PSNR, and LPIPS scores among the three variants. On
the Pix2Pix dataset, it outperforms wavelet-only conditioning on all
four metrics and matches, though does not surpass map-only
conditioning on LPIPS. FID is the sole exception to this pattern: it favors wavelet-only conditioning on both datasets, which comes out lowest of the three variants throughout. Combined conditioning doesn't close this gap: it trails both single-modality variants on our own data and lands between wavelet-only and map-only on Pix2Pix. This pattern echoes the perception-distortion tradeoff~\cite{blau2018perception}: tighter conditioning toward the paired target sharpens reference-based accuracy (SSIM, PSNR), but nothing guarantees it also sharpens distributional realism (FID). LPIPS tracking the reference-based metrics here is a feature of our results rather than something the tradeoff itself predicts, since LPIPS is usually grouped with FID on the perceptual side of that framework. Given how modest our test sets are (491 pairs for ours, 220 for Pix2Pix), and given FID's known bias at small sample sizes~\cite{chong2020effectively}, we treat this gap with some caution. That caution, together with FID's general limitations, is why we weight SSIM, PSNR, LPIPS, and visual inspection more heavily than FID in our conclusions. SSIM and PSNR favor wavelet and combined conditioning consistently across both datasets, with LPIPS mostly following suit.


The ablation study reinforces the same tradeoff along a different axis:
conditioning scale rather than modality choice. SSIM and PSNR are
highest at full strength (1.0/1.0) on both datasets, with the
wave-favoring setting (0.5/1.0) outperforming the map-favoring
setting (1.0/0.5) on PSNR in both datasets and on SSIM on our own
dataset, though the two settings are effectively tied on Pix2Pix
SSIM (0.160 vs.\ 0.162). FID shows the reverse ordering on both
datasets: (0.5/1.0) is lowest, (1.0/1.0) is highest, and (1.0/0.5)
falls in between. LPIPS is less consistent. On Pix2Pix, (1.0/1.0)
gives the worst LPIPS(VGG) and ties for the worst LPIPS(Alex). On
our dataset, (1.0/1.0) is slightly worse than (0.5/1.0) on
LPIPS(Alex) but is the best setting on LPIPS(VGG). Visual inspection
tracks the FID pattern: (0.5/1.0) looks most realistic on both test
sets, while (1.0/1.0) shows visible degradation alongside its higher
FID.

\subsection{Qualitative Results}
\label{fig:visual}
\begin{figure}[!htbp]
\begin{minipage}[t]{0.48\linewidth}
\centering
\renewcommand{\arraystretch}{0.5}
\setlength{\tabcolsep}{2pt}
\def\imgwEx{0.30\linewidth}


\begin{tabular}{ccc}
\textbf{\scriptsize OSM} & \textbf{\scriptsize Predicted} & \textbf{\scriptsize Ground Truth} \\[4pt]
\includegraphics[width=\imgwEx]{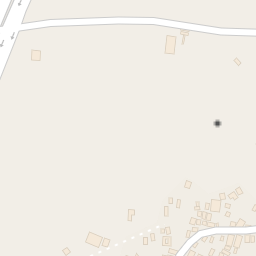} & \includegraphics[width=\imgwEx]{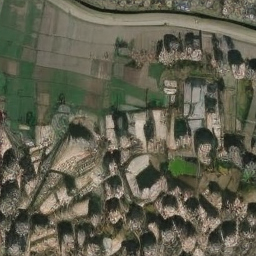} & \includegraphics[width=\imgwEx]{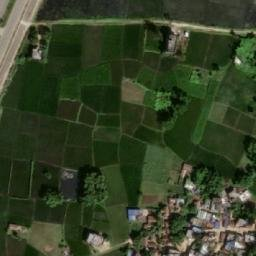} \\[4pt]
\includegraphics[width=\imgwEx]{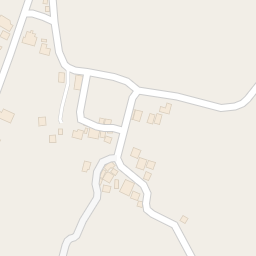} & \includegraphics[width=\imgwEx]{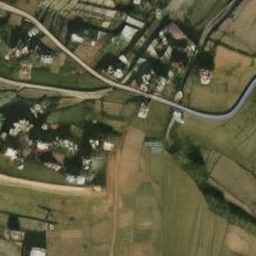} & \includegraphics[width=\imgwEx]{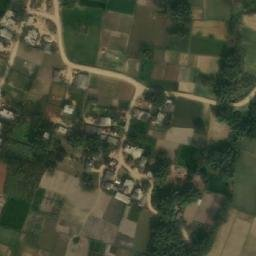} \\[4pt]
\end{tabular}
\captionof{figure}{Satellite image synthesis conditioned on OSM only. Left to
right: OSM input, prediction, ground truth.}
\label{fig:osm_examples}
\end{minipage}
\hfill
\begin{minipage}[t]{0.48\linewidth}
\centering
\renewcommand{\arraystretch}{0.5}
\setlength{\tabcolsep}{2pt}
\def\imgwEx{0.30\linewidth}

\begin{tabular}{ccc}
\textbf{\scriptsize Wavelet} & \textbf{\scriptsize Predicted} & \textbf{\scriptsize Ground Truth} \\[4pt]
\includegraphics[width=\imgwEx]{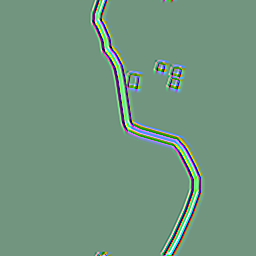} & \includegraphics[width=\imgwEx]{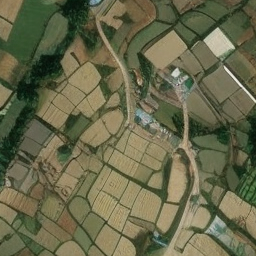} & \includegraphics[width=\imgwEx]{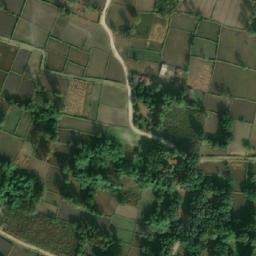} \\[4pt]
\includegraphics[width=\imgwEx]{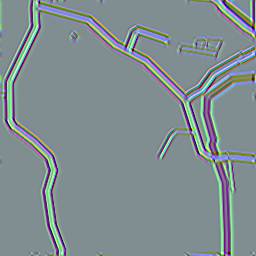} & \includegraphics[width=\imgwEx]{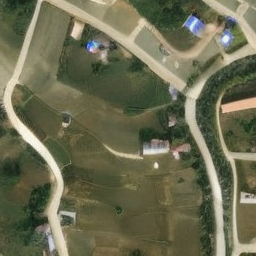} & \includegraphics[width=\imgwEx]{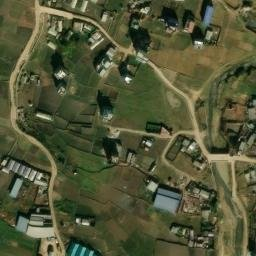} \\[4pt]
\end{tabular}
\caption{Satellite image synthesis conditioned on wavelet subbands of
the OSM image only. Left to right: wavelet subband input, prediction,
ground truth.}
\label{fig:wavelet_examples}
\end{minipage}
\end{figure}

\begin{figure*}[!htbp]
\centering

\newlength{\dividergap}
\setlength{\dividergap}{1em} 

\begin{minipage}[t]{0.45\linewidth}
\centering
\renewcommand{\arraystretch}{0.5}
\setlength{\tabcolsep}{2pt}
\def\imgw{0.22\linewidth}

\begin{tabular}{cccc}
\textbf{\scriptsize  OSM} & \textbf{\scriptsize Wavelet} & \textbf{\scriptsize OSM + Wavelet} & \textbf{\scriptsize  Ground Truth} \\[4pt]
\includegraphics[width=\imgw]{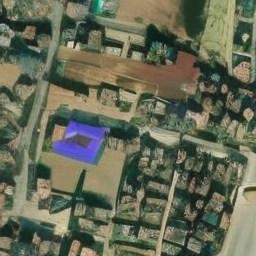} & \includegraphics[width=\imgw]{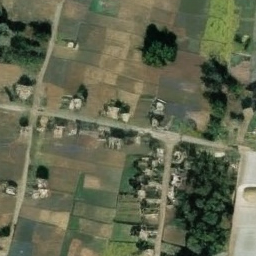} & \includegraphics[width=\imgw]{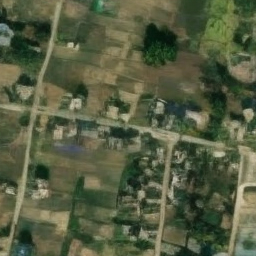} & \includegraphics[width=\imgw]{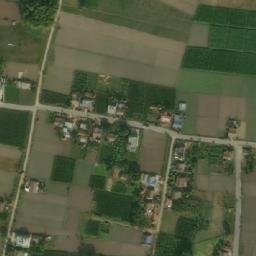} \\[4pt]
\includegraphics[width=\imgw]{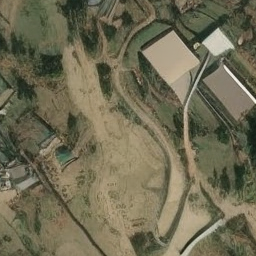} & \includegraphics[width=\imgw]{figures/wavelets/14.png} & \includegraphics[width=\imgw]{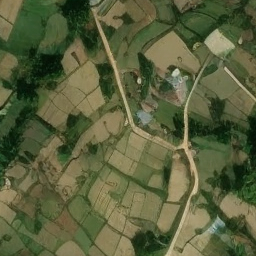} & \includegraphics[width=\imgw]{figures/gt/14.png} \\[4pt]
\includegraphics[width=\imgw]{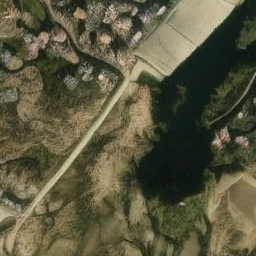} & \includegraphics[width=\imgw]{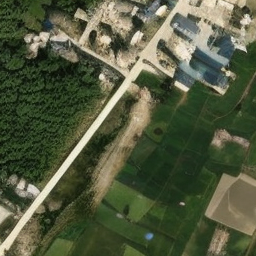} & \includegraphics[width=\imgw]{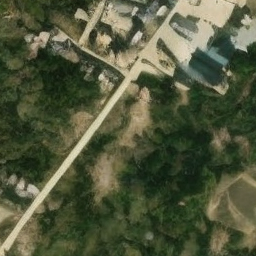} & \includegraphics[width=\imgw]{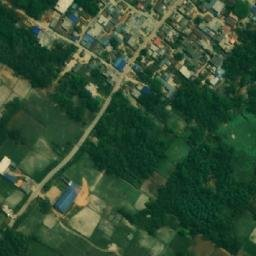} \\[4pt]
\includegraphics[width=\imgw]{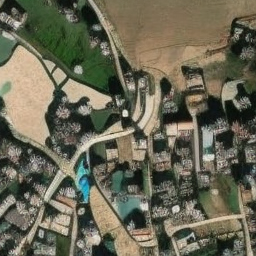} & \includegraphics[width=\imgw]{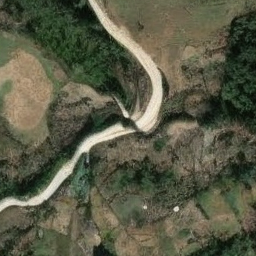} & \includegraphics[width=\imgw]{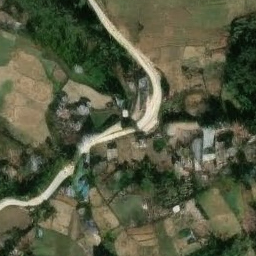} & \includegraphics[width=\imgw]{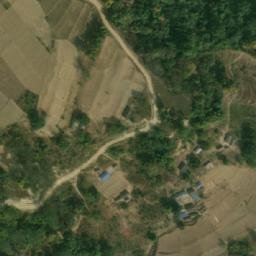} \\[4pt]
\end{tabular}
\end{minipage}%
\hspace{\dividergap}%
\vrule width 0.4pt%
\hspace{\dividergap}%
\begin{minipage}[t]{0.45\linewidth}
\centering
\renewcommand{\arraystretch}{0.5}
\setlength{\tabcolsep}{2pt}
\def\imgw{0.22\linewidth}

\begin{tabular}{cccc}
\textbf{\scriptsize OSM} & \textbf{\scriptsize   Wavelet} & \textbf{\scriptsize  OSM + Wavelet} & \textbf{\scriptsize  Ground Truth} \\[4pt]
\includegraphics[width=\imgw]{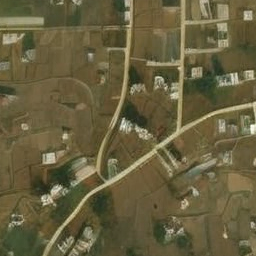} & \includegraphics[width=\imgw]{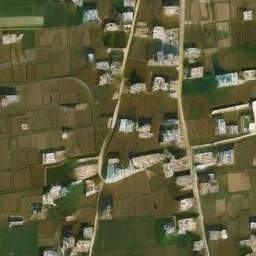} & \includegraphics[width=\imgw]{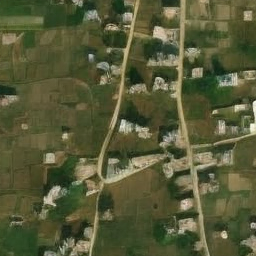} & \includegraphics[width=\imgw]{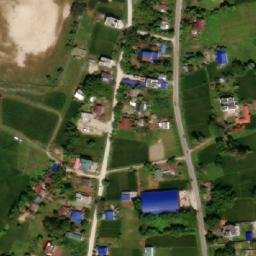} \\[4pt]
\includegraphics[width=\imgw]{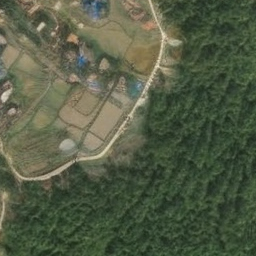} & \includegraphics[width=\imgw]{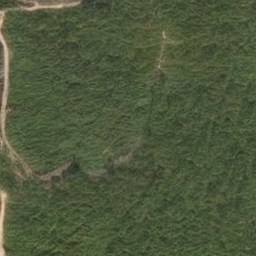} & \includegraphics[width=\imgw]{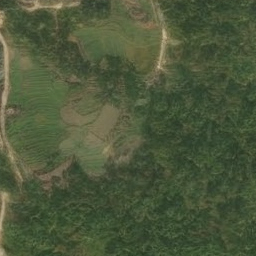} & \includegraphics[width=\imgw]{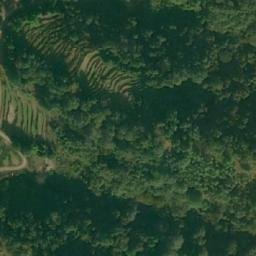} \\[4pt]
\includegraphics[width=\imgw]{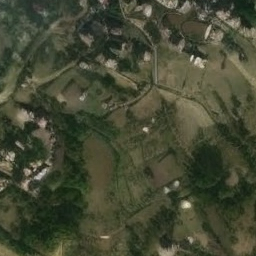} & \includegraphics[width=\imgw]{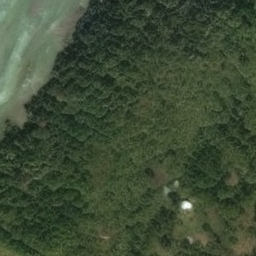} & \includegraphics[width=\imgw]{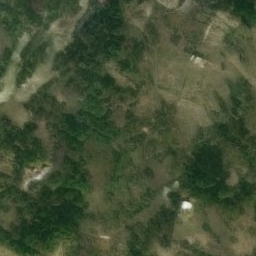} & \includegraphics[width=\imgw]{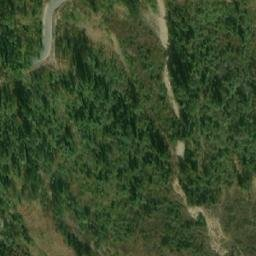} \\[4pt]
\includegraphics[width=\imgw]{figures/osm/2.png} & \includegraphics[width=\imgw]{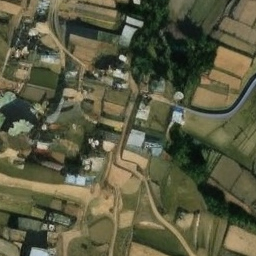} & \includegraphics[width=\imgw]{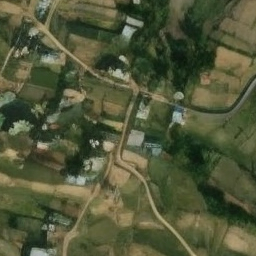} & \includegraphics[width=\imgw]{figures/gt/2.png} \\[4pt]
\end{tabular}
\end{minipage}

\caption{Qualitative comparison across the three conditioning variants.
From left to right: OSM-only conditioning, wavelet-only conditioning,
combined OSM + wavelet conditioning, and the ground-truth satellite
image.}
\label{fig:qualitative_4ex}
\end{figure*}

Under OSM-only conditioning (Fig.~\ref{fig:osm_examples}), road structure is
often missing/incomplete from the generated imagery, and the adapter does not
appear to have reliably learned terrain layout or plot boundaries from
the map raster during training. Wavelet-only conditioning (Fig.~\ref{fig:wavelet_examples}) tells a
different story, despite being derived from the same OSM tile, roads
are captured more reliably, and the adapter appears to have learned
terrain and plot layout more effectively, producing outputs closer to
ground truth overall although hallucinated building
footprints, with structures appear in the generated image at
locations where the ground truth contains none in all 3 variants

Combined conditioning also pushes road structure a step further than
wavelet-only already managed on its own, with road boundaries reading
as more continuous and better defined against the surrounding terrain.
A separate difference shows up in land-cover color: plotted
agricultural land that leans brownish under wavelet-only conditioning
shifts toward the greener tones actually present in the ground truth
once both signals are fused. We flag this purely as something noticed
on inspection and not a measured result since none of our reported metrics
isolate color accuracy on their own.

%% file: conclusion.tex
This work demonstrates that wavelet subbands derived from OpenStreetMap rasters provide a substantive conditioning signal to the the frozen Stable Diffusion backbone for map-to-satellite diffusion synthesis, rather than a marginal supplement to spatial-domain conditioning. The effect is most apparent in road network reconstruction: both wavelet-only and combined conditioning recover road structure with markedly higher fidelity than map-only conditioning, a result consistent with the directional subbands (LH, HL, HH) encoding linear, edge-like structure that a flat raster represents only implicitly. Qualitative inspection also shows frequency-decomposed conditioning identifies terrain layout and plot boundaries more reliably than map-only conditioning, whereas the map-only adapter frequently fails to recover these structures from the OSM raster alone. Taken together, these findings position frequency-aware conditioning as a promising direction for cartographic-to-satellite synthesis beyond the specific architecture examined here.

This framework also addresses a limitation present in several existing ControlNet-based methods~\cite{espinosa2023generate,sastry2024geosynth,tang2024crsdiff}, which condition on structural signals extracted directly from the target satellite image, such as canny, edge maps or segmentation masks. While effective when such imagery is already available, this dependence limits applicability precisely in the settings where synthesis offers the greatest value. Our framework instead conditions entirely on inputs obtainable independently of the target image, making it applicable to low-resource, imagery-scarce regions such as Nepal, where recent satellite coverage is limited and target-derived conditioning cannot be constructed.

\subsection{Limitations and Future Work}
Despite promising results, this work has several limitations. The
dataset, while diverse, is limited to 4,885 pairs and may not fully
capture the complete geographic variability of Nepal. Future work
will explore scaling the dataset and extending the framework to
additional South Asian geographic domains, which would also clarify
whether the road-structure gains attributable to wavelet conditioning
generalize beyond the terrain and OpenStreetMap tagging conventions
specific to Nepal. Road-specific evaluation metrics beyond SSIM and
PSNR could also be used to better quantify structural fidelity in
generated satellite imagery.